\documentclass[]{interact}

\usepackage{epstopdf}
\usepackage[caption=false]{subfig}

\usepackage[longnamesfirst,sort]{natbib}
\bibpunct[, ]{(}{)}{;}{a}{,}{,}
\renewcommand\bibfont{\fontsize{10}{12}\selectfont}

\theoremstyle{plain}

\theoremstyle{definition}

\theoremstyle{remark}

\begin{document}


\title{A survey of automatic de-identification of longitudinal clinical narratives.}

\author{
\name{Vithya Yogarajan\textsuperscript{a}\thanks{CONTACT Vithya Yogarajan. Email: vyogaraj@waikato.ac.nz}, Michael Mayo\textsuperscript{a} and Bernhard Pfahringer\textsuperscript{a}}
\affil{\textsuperscript{a}Department of Computer Science, The University of Waikato, Hamilton, New Zealand}
}

\maketitle

\begin{abstract}
Use of medical data, also known as electronic health records, in research helps develop and advance medical science. However, protecting patient confidentiality and identity while using medical data for analysis is crucial. Medical data can be in the form of tabular structures (i.e. tables), free-form narratives, and images. This study focuses on medical data in the free form longitudinal text.   De-identification of electronic health records provides the opportunity to use such data for research without it affecting patient privacy, and avoids the need for individual patient consent. In recent years there is increasing interest in developing an accurate, robust and adaptable automatic de-identification system for electronic health records. This is mainly due to the dilemma between the availability of an abundance of health data, and the inability to use such data in research due to legal and ethical restrictions. De-identification tracks in competitions such as the 2014 i2b2 UTHealth and the 2016 CEGS N-GRID shared tasks have provided a great platform to advance this area. The primary reasons for this include the open source nature of the dataset and the fact that raw psychiatric data were used for 2016 competitions. This study focuses on noticeable trend changes in the techniques used in the development of automatic de-identification for longitudinal clinical narratives. More specifically, the shift from using conditional random fields (CRF) based systems only or rules (regular expressions, dictionary or combinations) based systems only, to hybrid models (combining CRF and rules), and more recently to deep learning based systems. We review the literature and results that arose from the 2014 and the 2016 competitions and discuss the outcomes of these systems. We also provide a list of research questions that emerged from this survey. 
\end{abstract}


\section{Introduction}

Health records are an extremely valuable but arguably underutilised source of health information for researchers. Such research is made possible by the widespread adaptation of electronic health records (EHR) in routine health care (\cite{murdoch2013inevitable}). These records create large and continuously updated datasets which can then be subjected to powerful analytic techniques. 

De-identification of electronic health records (EHR) is the process of identifying and removing any information that will reveal patient identity from medical records. It is a method for protecting patient confidentiality and privacy before the use of EHR for research purposes. This avoids the need for individual patient consent before medical records (text or otherwise) are used in research. An automatic de-identification system allows researchers to use all the medical data sources available and utilise data-driven models to improve health outcomes. 

\begin{figure*}
\begin{center}
\includegraphics[width=0.75\textwidth]{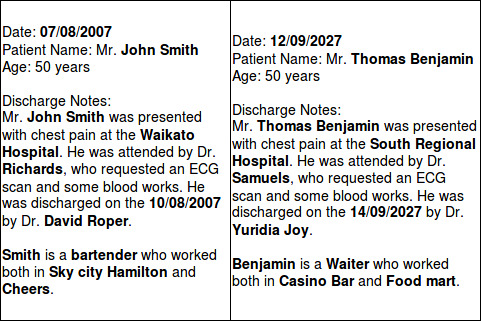}
\end{center}
\caption{Example of EHR before and after de-identification. }
\label{fig:surragate}
\end{figure*}

The de-identification of EHR consists of two stages. The first stage is to identify protected health information (PHI). The second stage is to replace the identified PHI with realistic surrogates. Figure \ref{fig:surragate} provides an example of EHR before and after de-identification. From this example it is evident PHI such as hospital name, patient name, date etc. are replaced with realistic surrogates. As illustrated in Figure \ref{fig:surragate}, after de-identification the structure and readability of the original text document should still be maintained (\cite{Stubbs2015}). 

One of the significant difficulties in de-identifying the records is complying with the legal requirements. Rules and regulations are provided by the Health Insurance Portability Accountability Act (HIPAA) to de-identify EHR (\cite{Stubbs2015,NIST}). In New Zealand, patient privacy is protected by the Health Information Privacy Code and the Health and Disability Consumers' Code of Rights (\cite{healthprivacy,healthdisability}). Although there is no one standard set of rules and methods for de-identification, the HIPAA Privacy Rule is arguably the worldwide gold standard (\cite{NIST}). HIPPA considers Expert Determination and Safe Harbor to be the two acceptable methods of de-identification. Expert Determination relies on a person with specialist knowledge and experience to ensure only a minimal, statistically acceptable level of risk of identifying an individual is present (\cite{NIST}). Organisations such as the Institute of Medicine (IOM), the Health Information Trust Alliance (HITRUST), the Pharmaceutical Users Software Exchange (PhUSE) and the Council of Canadian Academics, have adapted expert determination standards for de-identification (\cite{iapp}). 

The Safe Harbour method provides guidelines on 18 categories of PHI and requires all information that identifies the patient be to be removed before records are shared outside the clinical setting (\cite{STUBBS2015S20,psychatric}). These categories are presented in Table~\ref{table:HIPAA}.

\begin{table*}
\small\sf\centering
\caption{HIPAA categories for de-identification using the Safe Harbour method.}\label{table:HIPAA}
\centering
\begin{tabular}{ l l }
\toprule
{Rule } & {PHI Identifier Descriptions} \\
{Number} & \\
\midrule
{1}  & Names \\
{2} &  All geographic subdivisions smaller than state \\
 & (e.g. zip code, street address, city) \\
{3} & All elements of date (except year) \\
{4} & Telephone numbers \\
{5} & Fax numbers \\
{6} & Electronic mail addresses \\
{7} & Social security numbers\\
{8} & Medical record numbers\\
{9} & Health plan beneficiary numbers\\
{10} & Account numbers\\
{11} & Driver's certificate/license numbers\\
{12} & Vehicle identification and serial numbers, including license plate numbers\\
{13} & Device identifier and serial numbers\\
{14} & Web Universal Resource Locators (URLs)\\
{15} & Internet Protocol (IP) address numbers\\
{16} & Biometric identifiers\\
{17} & Full face photographic images and any comparable images\\
{18} & Any other unique identifying numbers, characteristics or codes.\\ 
\bottomrule
\end{tabular}
\end{table*}

The primary challenge for de-identification of unstructured clinical data is that while identifying and replacing PHIs the integrity of data needs to be protected (\cite{Meystre2010,10.2307/23216664,doi:10.1197/jamia.M2444}). This needs to be achieved in the presence of ambiguities (such as lexical overlap between PHI and non-PHI) and out-of-vocabulary words (such as misspelt and foreign words in PHIs) (\cite{doi:10.1197/jamia.M2444}). For example, while the de-identifying name, it is essential to recognise the difference between the patient name, doctor's name and a spouse (or family member) name that may also be in the EHR. There is a need to be consistent while generating surrogates. This issue is even more complicated when the data is longitudinal of nature. 

Another major issue is that studies have shown that even with very few entities (such as gender, date of birth, ZIP code, and address over a time) reconstructions of identity is possible (\cite{rothstein2010deidentification}). For example, Governor Weld of Massachusetts had his health records identified from ``anonymised'' records of state employee hospital visits (\cite{rothstein2010deidentification}). Patients are concerned about the privacy and confidentiality of their health information if it is used for research, especially concerning sensitive items such as sexual history, genetic disease and mental health (\cite{king2012perspectives}). However, these concerns decrease when demonstrable measures are taken to protect privacy (\cite{king2012perspectives}).

De-identification of EHR can be done manually or automatically. Manual de-identification is time consuming, tedious, costly, can only be done by a restricted set of individuals allowed access to the original patient notes, and is subject to human error (\cite{HE2015S39, DEHGHAN2015S53, doi:10.1093/jamia/ocw156,BUI2017}). There have been many attempts to develop an automatic system that will de-identify EHR with certainty. Several systems have been proposed, but they are not frequently used in practice. These include Anonym software (\cite{DBLP:journals/artmed/ZucconKNB14}), MITRE's MIST tool (\cite{ABERDEEN2010849}), CCHMC (\cite{CCHMC}), CRATE (\cite{CRATE}) and FASDIM (\cite{CHAZARD2014303}). In addition, Meystre \textit{et al.}, (2010) (\cite{Meystre2010}) presents an overview of several other de-identification systems that are available.  

Techniques used for de-identification of EHR fit into three categories: rule-based, machine learning based and hybrid, which is a combination of both rule-based and machine learning-based. Rule-based techniques use dictionaries and hand-coded rules, and rely on patterns expressed as regular expressions that are defined and tuned by humans (\cite{doi:10.1093/jamia/ocw156,BUI2017,psych1}). Rule-based systems also require an expert to provide knowledge of these rules. They do not need significant training data and are easily modifiable. Compared to machine learning approaches, rule-based approaches can distinguish some ambiguous instances, such as the difference between medical record numbers and phone numbers (\cite{LIU2015S47}). However, rule-based methods have limitations, as they require additional data curation or annotation by domain experts (\cite{psych1,YANG2015S30}). Also, with rule-based systems, the rules in one system are not easily adoptable by another system (\cite{LI2017}). Machine learning approaches, on the other hand, can be trained to recognise PHI patterns automatically. The most popular machine learning techniques used are conditional random fields (CRF)  (\cite{HE2015S39,DEHGHAN2015S53,BUI2017,LIU2015S47, YANG2015S30}), support vector machines (\cite{STUBBS2015S67, KHALIFA2015S128}), hidden Markov models (\cite{CHEN2015S60}),  and neural networks (\cite{doi:10.1093/jamia/ocw156}). However, the main downside to machine learning techniques is that they require significantly large labelled datasets as well as feature engineering (\cite{doi:10.1093/jamia/ocw156, YANG2015S30}). Machine learning systems perform worse than rule-based systems on rare types of PHI due to a dearth of training data (\cite{psych1}).
Hybrid systems, combining rules and machine learning, are currently the most popular and most effective systems (\cite{DEHGHAN2015S53,psych1,LIU2015S47,YANG2015S30,LI2017,STUBBS2015S67,Stubbs2017}).

An initial impression is that de-identification of EHR is a relatively straightforward problem. The basic idea of de-identification is simple, and as mentioned above there are many tools and software programs available. However, in reality, it is an essential and currently unsolved problem in the medical informatics domain. The main reason is that the accuracy, precision, robustness and adaptability of the available open source tools and systems are mediocre (\cite{ZHENG2015S189,Meystre2010,MEYSTRE2014142,HE2015S39, naturallang, STUBBS2015S67, STUBBS2015S78}). Examples of issues that complicate developing a well-performing de-identification system that adapts to change include the nature, source, and type of dataset; the time frame of the dataset; the intensity of PHI in the dataset; the kind of PHI; free-form text vs formally defined texts. Unfortunately, most systems -- BoB, MIT, MIST, HIDE, HMS, MeDS -- can misidentify clinical information as PHI (\cite{MEYSTRE2014142}). Ambiguous medical terms require more accurate de-identification techniques (\cite{MEYSTRE2014142}). Also, most of the open source software only target specific PHI and not all of them (\cite{Meystre2010}). It is argued that perfection cannot be achieved; however, 95\% accuracy is considered the rule of thumb and an acceptable value for accuracy (\cite{psychatric,STUBBS2015S11}).

In the past five years, there have been more than 100 papers published relating to the de-identification problem. One of the main reasons for the recent development in the literature is the EHR de-identification competitions. They provide an excellent platform to achieve better results, and the data is open source (\cite{STUBBS2015S20, STUBBS2015S11, BUI2017, Stubbs2017, psychatric}). There have been three main competitions relating to de-identification of EHR: the 2006 Informatics for Integrating Biology and the Bedside (i2b2) competition (\cite{KUMAR2015S6, naturallang}); the 2014 i2b2/UTHealth shared task (\cite{HE2015S39, naturallang, STUBBS2015S67, STUBBS2015S78}); and the 2016 Centers of Excellence in Genomic Science (CEGS) and Neuropsychiatric Genome-Scale and RDOC Individualized Domain (N-GRID) shared task (\cite{BUI2017, Stubbs2017, psychatric, Liu2017}). These competitions also included multiple tracks focusing on issues such as smoking status classification (in 2006); coronary artery disease (CAD) classification for diabetic patients (in 2014); classification of specific symptom severity based on the psychiatric evaluation (in 2016).  

Due to the abundance of literature in the recent years, this study restricts itself to the most recent work and provides an overview of the 2014 i2b2/UTHealth shared task, and 2016 CEGS N-GRID shared the task. This study primarily focuses on the trend changes noticed in the techniques used to develop de-identification systems. To the best of our knowledge, this paper is the most recent review paper since Meytre \textit{et al.}, (2010) (\cite{Meystre2010}).

The rest of the paper is structured such that an overview of the 2014 and 2016 competitions is presented in section 2, resulting de-identification systems in section 3, error analysis in section 4 and finally discussions in section 5.

\section{Overview of the recent Competitions}

This section provides a brief overview of the 2014 and 2016 competitions. Both 2014 and 2016 competitions used longitudinal data as it enables the understanding of each patient's medical records and their progress over time (months or years). It provides a more comprehensive clinical summary of patient-based clinical experience than encounter-based or provider-based data. However, the main issue with using longitudinal data is that it contains an abundance of personal information on each patient. Collectively, longitudinal data is more likely to allow the piecing together of patient identity (\cite{STUBBS2015S20,STUBBS2015S11}). Also, both competitions used ''as is'' raw data and reflects the realities of what the de-identification system would have to handle in reality (\cite{naturallang}). Data was split 60/40 and distributed into two parts. Initially, the training set for the development of the system was distributed followed by testing set on a later date. For more details on the 2014 competition see Kumar \textit{et al.}, (2015) (\cite{KUMAR2015S6}), Stubbs and Uzuner, (2015) (\cite{STUBBS2015S20}), Stubbs \textit{et al.}, (2015) (\cite{STUBBS2015S67}), Stubbs and Uzuner, (2015) (\cite{STUBBS2015S78} and Stubbs \textit{et al.}, (2015) (\cite{STUBBS2015S11}). Details of the 2016 competition is presented in:  Stubbs \textit{et al.}, (2017) (\cite{psychatric}), Bui \textit{et al.}, (2017) (\cite{BUI2017}) and Uzuner \textit{et al.}, (2017) (\cite{UZUNER2017S1}).

\subsection{2014 i2b2/UTHealth Shared task}
The 2014 i2b2/UTHealth shared task used diabetic patient longitudinal medical
records obtained from two large hospitals. The 2014 competition included four tracks. Track 1, the de-identification track, is the primary interest of this survey. Track 1 was attempted by ten teams who submitted 22 system outputs for evaluation. The best performing system of each team was evaluated, and three of the systems achieved F-measures over 0.90 (\cite{STUBBS2015S11}). A number of the top performing systems used a hybrid model combining regular expressions and rules with machine learning based systems (\cite{DEHGHAN2015S53,STUBBS2015S11, YANG2015S30}). The conditional random field was the most popular machine learning system. Categories such as ZIP, AGE and PHONE were considered easier, and PROFESSION and LOCATION categories proved to be more difficult (\cite{STUBBS2015S11}). It was no surprise that overall results for i2b2-PHI categories are worse than that of HIPAA-PHI categories, as i2b2-PHI had much more specific categories than HIPAA-PHI (\cite{HE2015S39,DEHGHAN2015S53,LIU2015S47, YANG2015S30}).

\subsection{The 2016 CEGS N-GRID Shared Task}

The 2016 CEGS N-GRID shared task used psychiatric data, making it the first ever competition to release psychiatric intake records (\cite{psychatric,psych1}). 
Psychiatric data much more challenging to de-identify, as they contain higher density of PHIs, and is one of the main reasons why mental health notes are not commonly used for research purposes (\cite{psychatric,BUI2017, TRAN2017S138,UZUNER2017S1}). 


The 2016 competition included three tracks, of which the first track, track 1A and 1B, is the focus here. Track 1A was designed to evaluate existing systems from 2014 on the 2016 data without any further training or modification. It was designed to test how well the systems performed on, and adapted to, data that is different from the system's original training data (\cite{psych1,UZUNER2017S1}). The best performing system on this data for track 1A was from the Harbin Institute of Technology Shenzhen Graduate School team, with an F-measure of 0.7985 (\cite{psychatric}). It is important to note that the fifth best team scored 0.5266 and seventh team 0.1861; therefore the spread was vast. In general, most of the existing systems submitted to this competition did not perform well on a new dataset (\cite{psychatric}). One potential reason for this is the nature of the dataset, as a psychiatric dataset has more personal information and more text. However, it is a good start at building models that can be tuned to new data (\cite{UZUNER2017S1}).

Track 1B was designed as a traditional evaluation system, and the systems were trained and developed on the psychiatric dataset (\cite{psych1}). The best performing system for track 1B was from the Harbin Institute of Technology Shenzhen Graduate School again team with an F-measure of 0.9143 (\cite{psychatric}). Seven of the top teams had an F-measure higher than 0.8 with the tenth best team having a score of 0.7175 (\cite{psychatric}). Nearly all teams used hybrid systems for track 1B. Different PHI categories were detected by different systems (rules or machine learning) and were combined at the end to address all PHI categories.  For example, some systems took the approach of using combinations of CRF addressing different PHI, and others combined CRF with bi-directional long-short-term memory systems (\cite{UZUNER2017S1}). The top-performing teams made use of neural networks (\cite{psychatric}). Teams found PHI categories PROFESSION, LOCATION and ID to be the most difficult categories, and DATE and AGE were the easiest to detect (\cite{psychatric}).

\subsection{Gold Standard}
The gold standard for 2014 and 2016 competitions training and testing corpus was developed using both manual and automatic processing (\cite{STUBBS2015S20,STUBBS2015S78}). Experts with special training were used as annotators who worked in parallel to each other. Data were de-identified under a risk-averse  interpretation of HIPAA guidelines (\cite{STUBBS2015S20}). This included the use of double annotation, arbitration, rounds of sanity checking and proofreading (\cite{STUBBS2015S20}). 
While preparing the gold standard, the semantics and integrity of the data were maintained such that the resulting data was useful and maximises the usability of the data for research. It was ensured that the gold standard contained adequate representations of PHI categories for the machine learning based system to learn automatically. Also, the gold standard was created by replacing PHIs of the raw data with realistic surrogates using both automatic systems and hand curation.

\subsection{Evaluation}\label{evaluation}

The NLP system, designed for the 2014 and 2016 competitions, contained seven main PHI categories and 25 associated sub-categories (\cite{DEHGHAN2015S53,YANG2015S30, naturallang}). The seven main categories were Name, Profession, Location, Contacts, IDs, Age and Dates (\cite{DEHGHAN2015S53}). Example of the sub-categories includes patient name, doctor's name and username for NAME category (see Tubbs and Uzuner, 2015 (\cite{STUBBS2015S20} for full detail of these categories). The use of sub-categories was only introduced in the 2014 competition, as the 2006 competition just focused on the seven types of PHI. 

It was argued that HIPAA only provides a starting point for an effective de-identification and that expanding the HIPAA definitions is believed to be more effective in solving the problem  (\cite{STUBBS2015S11}). This is the primary idea behind i2b2-PHI, where 25 subcategories were added to HIPAA-PHI to create i2b2-PHI categories (\cite{STUBBS2015S11}). In addition to the standard HIPAA-PHI, i2b2-PHI were also used in the competitions as an evaluation tool.  

De-identification performance is evaluated using Precision (P), Recall (R) and F-measure (F1), where, 
\begin{equation}
\begin{split}\label{eqn:evaluation}
P & = \frac{true \ positive \ rate}{true \ positive \ rate + false \ positive \ rate} \\
R & = \frac{true \ positive \ rate}{true \ positive \ rate + false \ negative \ rate} \\
F1 & = 2\times\frac{P \times R}{P+R} \\
\end{split}
\end{equation}

Evaluations of P, R and F1 were performed at the entity-level and token-level (\cite{LIU2015S47,YANG2015S30,STUBBS2015S67}). For entity-level, an exact match of the gold standard entity with the predicted entity is required. Whereas, for token-level, each token is considered separately. One token of the golden standard entity and the predicted entity is needed to match. Evaluations were performed against the gold standard of i2b2 test data (\cite{YANG2015S30}), using strict evaluation and relaxed evaluation. The strict metric involves exact string matching of the predicted entity against the gold standard. The relaxed metric, on the other hand, is approximate string matching with some leeway allowance at the end of a string. The overall performance of the system was measured using both micro and macro evaluation matrics (\cite{YANG2015S30}). Macro evaluation calculates the scores for each document and then takes the average across the corpus, and micro evaluation calculates the scores for each annotation across the corpus. This study will focus mainly on the micro-averaged F-measure as this was the primary evaluation measure used for determining the performances of the systems.

\section{Resulting de-identification Systems}

This section presents the de-identification systems developed using the 2014 dataset during and after the competitions. It also presents de-identification systems developed as part of the 2016 competitions. It is important to note that at the stage of this study, there was only evidence of the availability and use of the 2014 dataset, after the competitions.

\subsection{Top performing De-identification system in the 2014 competitions}

This section provides an overview of some of the best-performing systems developed during the 2014 competition, track 1 (de-identification track).

Table \ref{table:evaluation} presents overall performance micro F-measure for these systems. The micro F-measures are the primary evaluation method for the de-identification task (task 1) (\cite{STUBBS2015S11}). Overall, the HIPAA-PHI micro F-measure is always better than that of i2b2-PHI micro F-measure. This is mainly because i2b2-PHI categories include more PHI categories to de-identify than HIPAA-PHI categories. Three out of the top ten teams had an F-measure of above 0.9 for the i2b2:PHI categories and four out of the top ten teams had an F-measure of above 0.9 for the HIPAA:PHI categories (\cite{STUBBS2015S11}). Table \ref{table:bestworst} presents the best and worst performing PHI categories for these systems. 

\begin{table}[h!]
\small\sf\centering
\caption{The overall performance of some of the top teams in 2014 competitions. 2014 Top teams includes S1 (\cite{YANG2015S30}), S2 (\cite{LIU2015S47}), S3 (\cite{DEHGHAN2015S53}), S4 (\cite{HE2015S39}) and S5 (\cite{CHEN2015S60}).}
\label{table:evaluation}
\centering
\begin{tabular}{llll}
\toprule
\multicolumn{2}{l}{De-identification System} &  {i2b2-PHI} & {HIPAA-PHI} \\  
Number & Description & {Micro F-measure} &    {Micro F-measure} \\ 
 \midrule
S1&{Yang and Garibaldi (2015) } & 0.9360  & 0.9573 \\
S2&{Liu \textit{et al.}, (2015) } & 0.9124  & 0.9409\\ 
S3&{Dehghan \textit{et al.}, (2015) } & 0.9065  & 0.9323 \\ 
S4&{He \textit{et al.}, (2015) } & 0.8852  & 0.9180 \\ 
S5&{Chen \textit{et al.}, (2015)} & 0.7549 & 0.7985 \\ 
\bottomrule
\end{tabular}
\end{table}

\begin{table}[h!]
\small\sf\centering
\caption{F-measures of best and worst performing categories in 2014 de-identification systems.}
\label{table:bestworst}
\begin{center}
\begin{tabular}{lllll}
\toprule
{Systems} & {Best PHI} & {F1} & {Worst PHI} & {F1} \\ 
&{categories} & &{categories} & \\
\midrule
S1 & DATE, AGE, & 0.94-  & PROFESSION & 0.69 \\
  & CONTACT, NAME &0.97 & & \\
S2 & AGE, DATE & 0.94, & PROFESSION, & 0.69,0.79  \\ 
  & & 0.97  &  LOCATION &  \\
S3  & AGE, DATE, EMAIL, & $>$ 0.94 & PROFESSION, & 0.57, 0.27 \\
  &  PHONE, STREET, ZIP,  & & ORGANISATION & \\
 & MEDICAL, RECORD   & & & \\
S4  & USERNAME, DATE, & $>$ 0.97 & PROFESSION,  & 0.40- 0.69\\ 
 &  MEDICAL,  RECORD & & ORGANISATION,  &   \\
&  & & COUNTRY & \\
S5  & NAME, CONTACT & $>$ 0.91 & PROFESSION, AGE  & 0.42, 0.59  \\ 
\bottomrule
\end{tabular}
\end{center}
\end{table}

Table \ref{table:2014type} presents examples of techniques used by the top teams in 2014 competitions to de-identify PHIs. One of the main observation with 2014 competitions is that there was no one uniquely favouring technique to any given PHI. CRF was the primary machine learning technique used in the 2014 competitions, with one exception presented by Chen \textit{et al.}, (2015) (\cite{CHEN2015S60}). Chen \textit{et al.}, (2015) (\cite{CHEN2015S60}) presented a non-parametric Bayesian hidden Markov model using a Dirichlet process (HMM-DP) and was the only top team to present a machine learning only system to de-identify all PHI's. Handcrafted rules and regular expressions used in the 2014 competitions are particular to the dataset. Table \ref{table:reg2014} presents examples of regular expressions used in the 2014 competitions for both tokenization (as part of pre-processing) and PHI identification.   

\begin{table}[h!]
\small\sf
\begin{center}
\caption{Examples of combination of techniques used to de-identify PHI categories and sub-categories in 2014 competitions.}
\label{table:2014type}
\begin{tabular}{lll}
\toprule
PHI categories& Sub-categories & Techniques \\
\midrule
{DATE} & Date &  CRF + Rules + Dictionary (S4); CRF + Rules (S3); \\ & &  CRF + Rules + Keywords (S1); HMM-DP  (S5)   \\
{NAME} & Doctor &  CRF + Rules (S4); CRF + Keywords (S1); HMM-DP (S5)   \\
&    Patient &  CRF +Rules (S3, S4); CRF + Keywords (S1); HMM-DP (S5) \\
&    Username & Rules (S3); CRF + Rules (S1); HMM-DP (S5)\\
{AGE} &    Age &   CRF + Rules (S4);  CRF + Rules +  Keywords (S1); \\ & & Rules (S3); HMM-DP (S5) \\
{CONTACT} &  Phone & Rules (S2,S3); CRF + Rules +  Keywords (S1); HMM-DP (S5)   \\
&    Fax, Email & Rules (S1,S2,S3); HMM-DP (S5)  \\
&    URL & Keywords (S1) \\
{ID} &    Medicalrecords & Rules  (S2,S3); HMM-DP (S5); CRF + Rules + Keywords (S1) \\
&    IDNUM & Rules (S2,S3); CRF + Rules +  Keywords (S1); \\ & & Rules +  Keywords (S1); HMM-DP (S5) \\
&    Device & Rules + Keywords (S1); HMM-DP (S5)   \\
&    BioID & Rules (S1) \\
&    Healthplan & Keywords (S1) \\
{LOCATION} & Hospital & CRF + Dictionary (S3); CRF +  Keywords (S1); HMM-DP (S5)   \\
&    City &  CRF + Rules + Dictionary (S4); CRF + Dictionary (S2,S3);  \\ & & CRF +  Keywords (S1); HMM-DP (S5) \\
&    State &  CRF + Rules + Dictionary (S4); CRF + Dictionary (S2); CRF \\ & & + Keywords (S1); Rules + Dictionary (S3); HMM-DP (S5)  \\
&    Street &  CRF + Rules + Dictionary (S4); Rules (S3); HMM-DP (S5)  \\
&    Zip & Rules (S3); CRF + Dictionary  (S2); CRF + Rules (S1); \\ & &  HMM-DP (S5)  \\
&    Organisation & CRF + Dictionary (S3); CRF +  Keywords (S1); HMM-DP (S5)   \\
&    Country &  CRF + Rules + Dictionary (S4); CRF + Dictionary  (S2); \\ & &   CRF +  Keywords (S1); Rules + Dictionary (S3); HMM-DP (S5)  \\
&    Location-Others & Keywords (S1); HMM-DP (S5)  \\
{PROFESSION} &  Profession & CRF + Dictionary (S3); CRF +  Keywords (S1); HMM-DP (S5)  \\
\bottomrule
\end{tabular}
\end{center}
\end{table}

\begin{table}[h!]
\small\sf\centering
\caption{Examples of regular expressions used in 2014 and 2016 competitions for both tokenization and PHI identification. 
}  
\label{table:reg2014}
\begin{center}
\begin{tabular}{llll}
\toprule
{Processing} & {Remarks}   & {Regular expression} & {Examples}  \\
{Stage} & & & \\
\midrule
{2014 - PHI}  & PHONE   & \footnotesize{\verb!\(\d{3}\)[− \t]?\d{3}[− \t]\d{4}!} & (871) 720-9439    \\
{identification} & (S2) & \footnotesize{\verb!\d{3}[− \t]\d{3}[− \t]\d{4}!} & 171-289-0968    \\
& & \footnotesize{\verb!\d{3}j\d{1}−\d{4}!} & 659-5187 \\ 
{2016 - PHI}  & PHONE   &\footnotesize{\verb!\(?(\d{3}\)?[-\t]?\d{3}[-\t]\d{4}!} & (304) 911-4864     \\
{identification} & (S6)  & \footnotesize{\verb!\d{3}\.\d{3}\.\d{4}!} & 017.445.0833 \\
\hline
{Tokenization }  & Before: & \footnotesize{\verb!∖d{2}/∖d{2}/∖d{4}!} & After:  \\ 
(2014) (S4) & {PEND01/26/2098} & & {PEND 01/26/2098} \\
{Tokenization } & Before: & \footnotesize{\verb!\d{1,2}([/-])(\d{1,2}(\1))?\d{2,4}!} & After: \\
(2016) (S10) & {09/14/2067CPT} & & {09/14/2067␣CPT} \\
\bottomrule
\end{tabular}
\end{center}
\end{table}

Evidently, in the 2014 competition, many top systems preferred hybrid approaches with CRF being the most popular machine learning approach. These de-identification systems included data specific rules and regular expressions to tackle PHIs such as ID and CONTACT. Also, most teams performed well in PHI categories, such as NAME, AGE and DATE, and poorly in PROFESSION and LOCATION.

 \subsection{Top performing de-identification system in the 2016 competitions}
This section provides an overview of some of the best-performing systems in the 2016 competition, track 1B (de-identification track).

Table \ref{table:evaluation16} presents overall performance F-measure for these systems. 
As with 2014 results, HIPAA PHI results are better than that of N-GRID PHI categories. This is no surprise as N-GRID PHI includes all subcategories. Two out of the top ten teams had an F-measure of above 0.9 for the N-GRID PHI categories and three out of the top ten teams had an F-measure of above 0.9 for the HIPAA PHI categories (\cite{psychatric}). Table \ref{table:bestworst16} presents the best and worst performing PHI categories for these systems.

\begin{table}[h!]
\small\sf\centering
\caption{The overall performance of some of the top teams in 2016 competitions for Track 1B, F-measures for N-GRID entity and HIPAA entity.2016 Top teams includes S6 (\cite{LI2017}), S7 (\cite{psych1}),  S8 (\cite{data-driven}), S9 (\cite{BUI2017}), and S10 (\cite{JIANG2017S43})}
\label{table:evaluation16}
\centering
\begin{tabular}{llll}
\toprule
\multicolumn{2}{l}{De-identification System} &  {i2b2-PHI} & {HIPAA-PHI} \\  
Number & Description & {Micro F-measure} &    {Micro F-measure} \\ 
 \midrule
S6 & {Lui \textit{et al.}, (2017)}  & 0.9143  & 0.9293 \\
S7 & {Lee \textit{et al.}, (2017)}  & 0.9074 & 0.9247 \\ 
S8 & {Dehghan \textit{et al.}, (2017)}  & 0.8769 & 0.8993 \\ 
S9 & {Bui \textit{et al.}, (2017)}  & 0.8731 & 0.9102 \\ 
S10 & {Jiang \textit{et al.}, (2017)}  & 0.8570 & 0.8882 \\ 
\bottomrule
\end{tabular}
\end{table}

\begin{table}[h!]
\small\sf\centering
\caption{F-measures of best and worst performing categories in 2016 de-identification system.}
\label{table:bestworst16}
\centering
\begin{tabular}{lllll}
\toprule
{Systems}   & {Best PHI} & {F1} & {Worst PHI} & {F1} \\ 
&{categories} & &{categories} & \\
\midrule
S6  & NAME, DATE,  & 0.92 - 0.97  & PROFESSION, & 0.65 - 0.85  \\
 &CONTACT, AGEE & &LOCATION, ID   & \\
S7  & NAME, DATE,  & 0.92 - 0.96  & PROFESSION, & 0.73 - 0.85  \\ 
 &CONTACT, AGE  & & LOCATION, ID  & \\
S8  & AGE, DATE,  & 0.92 - 0.95 & ORGANISATION, & 0.56, 0.69 \\ 
 &DOCTOR & & PROFESSION & \\
S9  & DATE, AGE,  & 0.91 - 0.96 & PROFESSION, ID, & 0.64 - 0.77 \\ 
 & CONTACT& & LOCATION & \\
S10  & AGE, DOCTOR,  & 0.94 - 0.96 & ORGANISATION, ID, & 0.69 - 0.77 \\
 & CONTACT, DATE  & & PROFESSION, FAX & \\
\bottomrule
\end{tabular}
\end{table}

Table \ref{table:2016type} presents examples of techniques used by the top teams in 2016 competitions to de-identify PHIs. In contrast to the 2014 competition, there was a noticeable shift towards the use of deep learning approaches such as bidirectional long short-term memory network (bi-LSTM). Hybrid systems were the preferred choice by the top performing teams. PHIs such as PHONE, FAX, EMAIL, URL and LICENSE only occur very rarely; hence hand-crafted rules only are used to identify (\cite{JIANG2017S43,LI2017,psych1}). Table \ref{table:reg2014} presents examples of regular expressions used in the 2016 competitions for tokenization and PHI identification. Due to the nature of the dataset, teams did find de-identification of 2016 dataset more difficult than that of 2014.

\begin{table}[h!]
\small\sf\centering
\begin{center}
\caption{Examples of combination of techniques used to de-identify PHI categories and sub-categories in 2016 competitions.}
\label{table:2016type}
\centering
\begin{tabular}{lll}
\toprule
PHI categories& Sub-categories & Techniques \\
\midrule
{DATE} & Date & CRF + Rules (S7,S9); CRF + Dic (S8); \\ & & Bi-LSTM (S6); LSTM (S10)  \\
{NAME} & Doctor & CRF (S7); CRF + Dic + Rules  (S9); LSTM (S10);    \\ & & CRF + Dic (S8); Bi-LSTM (S6)  \\
&    Patient &  CRF (S7); CRF + Dic + Rules  (S9);  LSTM (S10); \\ & & CRF + Dic (S8); Bi-LSTM (S6)  \\
&    Username & CRF + Rules (S7); Rules  (S9); Bi-LSTM (S6); LSTM (S10) \\
{AGE} &    Age &  CRF +  Rules (S7,S9); CRF + Dic (S8); \\ && Bi-LSTM (S6); LSTM (S10)  \\
{CONTACT} & Phone & CRF + Rules (S7); Rules  (S6,S9); Dic (S8); LSTM (S10)  \\
&    Fax & CRF (S7); Rules (S6,S9); Dic (S8); LSTM (S10)  \\
&    Email, URL & CRF + Rules (S7); Rules (S6,S9); Dic (S8); LSTM (S10)  \\
{ID} &    Medicalrecords & Rules (S7,S9); Bi-LSTM (S6); LSTM (S10) \\ 
& Healthplan & LSTM (S10) \\
& Licence & Rules (S6,S7,S9); Dic (S8); LSTM (S10) \\
&    IDNUM & Rules (S7,S9); Dic (S8); Bi-LSTM (S6); LSTM (S10)  \\
{LOCATION} & Hospital &  CRF (S7); CRF + Dic + Rules  (S9); LSTM (S10); \\ & & CRF + Dic (S8); Bi-LSTM (S6)   \\
&    City &  CRF (S7); CRF + Dic  (S8,S9); Bi-LSTM (S6);  LSTM (S10)   \\
&    State & CRF + Rules (S7); CRF + Dic  (S9); Dic (S8);  \\ & & Bi-LSTM (S6); LSTM (S10)  \\
&    Street & CRF + Rules (S7); CRF + Dic + Rules  (S9); Dic (S8); \\ & &  Bi-LSTM (S6); LSTM (S10)  \\
&    Zip &  CRF (S7); Dic (S8); Bi-LSTM (S6); LSTM (S10)   \\
&    Organisation &  CRF (S7); CRF + Dic + Rules (S9); LSTM (S10); \\ & &  CRF + Dic (S8); Bi-LSTM (S6)     \\
&    Country &  CRF (S7); CRF + Dic  (S8,S9); Bi-LSTM (S6);  LSTM (S10)   \\
&    Location-Others &   CRF (S7,S9); Bi-LSTM (S6); LSTM (S10)  \\
{PROFESSION} &    Profession &  CRF (S7); CRF + Dic  (S8,S9); Bi-LSTM (S6); LSTM (S10) \\
\bottomrule
 \end{tabular}
\end{center}
\end{table}

\subsection{Recent development using Competitions Data} 

This section presents development and improvement in automatic de-identification of EHR using the 2014 competition datasets. At the stage of this study, there was only evidence of the availability and use of 2014 competition dataset. The 2016 competition data is not available as an open source.  There are many examples --such as Lee \textit{et al.}, (2017) (\cite{DBLP:journals/corr/LeeDS17a}), Dernoncourt  \textit{et al.}, (2017) (\cite{DBLP:journals/corr/DernoncourtLS17}), Cormack \textit{et al.}, (2015) (\cite{CORMACK2015S120}),  Dai \textit{et al.}, (2015) (\cite{dai2015}-- where the 2014 dataset is used for experiments. This paper will focus only on the work relating to de-identification systems (similar to that of track 1). 

Table \ref{table:2014typeafter} presents examples of techniques used by systems developed using 2014 dataset to de-identify PHIs. In comparison to Table \ref{table:2014type} there is a definite shift to using deep learning methods such as LSTM and recurrent neural networks (RNN). Although the expected 95\% accuracy was not achieved in all PHIs by these systems, the overall F-measure was above 0.9. Also, all systems reported better F-measures to that of the 2014 competitions systems for poor performing PHIs such as PROFESSION and LOCATION. The change in trend, using deep learning only systems are welcoming and will be of interest to explore further.

\begin{table}[h!]
\small\sf\centering
\begin{center}
\caption{Techniques used to de-identify PHI categories in systems developed after the competitions using the 2014 dataset. These systems includes S11 (\cite{Phuong2016AutomaticDO}), S12 (\cite{doi:10.1093/jamia/ocw156}), S13 (\cite{kumar2016}) and S14 (\cite{Li2016}).}
\label{table:2014typeafter}
\centering
\begin{tabular}{lll}
\toprule
{PHI categories} & {Sub-categories} & {Techniques} \\
\midrule
{DATE} & Date & CRF + Rules (S11); LSTM (S12); RNN (S13, S14) \\ 
{NAME} &  Doctor, Patient & CRF + Rules (S11); LSTM (S12); RNN (S13, S14)  \\
&    Username & CRF (S11); LSTM (S12); RNN (S13, S14)  \\ 
{AGE} &    Age & CRF + Rules (S11); LSTM (S12);  RNN (S13, S14)  \\ 
{CONTACT} &    Phone & CRF + Rules (S11); LSTM (S12); RNN (S13, S14)  \\
&    Fax, Email, URL & CRF (S11); LSTM (S12);  RNN (S13, S14)  \\
{ID} &   all sub-categories  &  CRF (S11); LSTM (S12); RNN (S13, S140 \\
{LOCATION} & all sub-categories & CRF + Rules (S11); LSTM (S12);  RNN (S13, S14)  \\
{PROFESSION} &  Profession & CRF (S11); LSTM (S12); RNN (S13, S14) \\
\bottomrule
 \end{tabular}
\end{center}
\end{table} 

\section{Error Analysis}
 
Several of the de-identification systems developed during the competitions performed independent error analysis. Table \ref{table:erroranalysis} describes the most common types of errors that were defined by both 2014 and 2016 competition participants (\cite{HE2015S39, DEHGHAN2015S53,  YANG2015S30}). 

\begin{table}
\small\sf\centering
\caption{Definitions of common types of errors.}
\label{table:erroranalysis}
\centering
\begin{tabular}{ll}
\toprule
{Error type} &  {Description} \\
\midrule
 {Label errors} & An entity that belongs to category X is wrongly classified into \\ & category Y  
 or entity identified by correct start and end string but \\ &  wrong type. For example, ``John Smith'' classified as location. \\ 
& \\ 
 {Extent error} & The location span of the entity overlapped  with that of a gold \\ & standard but  not matched exactly. This could mean an entity has \\ & an additional or missing part. For example, in ``John Smith'' \\ & only ``Smith'' is classified as name. \\ 
& \\ 
 {Spurious errors} &  An entity is found in the system but not in the gold standard. \\ & This means the location span of the entity had no overlap with \\ & any correct entity. For example, ``visited the hospital'' \\ &  classified as name. \\ 
& \\ 
 {Missing error} & The location span of the entity in the gold standard had no  \\
 {(also called as}  & overlap with that of any entity in the system. This also refers   \\ 
 {omissions errors)} & to an entity found in the gold standard but not in the system  \\
& output. For example, ``John Smith'' not classified as a PHI. \\ 
& \\ 
\bottomrule
\end{tabular}
\end{table}

In both 2014 and 2016 competitions, most of the label errors fall into inter-PHI ambiguity instances, such DOCTOR and PATIENT. It is challenging to distinguish between PATIENT and DOCTOR as they both fall under the same main category, NAME (\cite{HE2015S39, YANG2015S30}). The 2016 competitions found sub-categories LOCATION and CONTACT had similar problems (\cite{JIANG2017S43}). The main issue with a label error is it will leave false placeholders and can cause confusion if it is a false positive (\cite{psych1}). If it is a false negative it is missed entirely.

In the 2014 competition, it was observed that among the PHI categories DATE is responsible for one-third of the extent errors. Yang and Garibaldi (2015) (\cite{YANG2015S30}) reported that out of the 1161 errors 69.7\% occurred due to missing errors. He \textit{et al.}, (2015) (\cite{HE2015S39}) also reported missing errors to be 64.42\%. In 2016 competitions, missing error occurs more often in subcategories with few instances of training data. For example, EMAIL, URL, MEDICAL RECORD; CONTACT, PROFESSION and ORGANISATION had the highest missing rates  (\cite{JIANG2017S43}).  Missing errors are also more serious as they have the highest risk of revealing patient identity (\cite{HE2015S39,psych1}). 

In comparison to missing errors, the proportion of spurious errors is much lower (\cite{HE2015S39}). 
Spurious errors occurred when tokens had similar lexical or contextual features with real PHI entities - e.g., DATE, AGE (\cite{JIANG2017S43}). It falsely removes non-PHI information which may be important for clinical or research applications (\cite{psych1}). 

There were additional sources of error that were mentioned concerning the 2016 competition in comparison to that of 2014. These include, acronyms and misspelt words, unknown name entities and errors due to the corpora and annotation problems. Acronyms and misspelt words are caused by wrongly recognising and classifying PHIs. However, recognising and classifying acronyms and abbreviations is hard as they can be medical terms or non-medical terms. For example, NAME: NC, NG etc. and DATE: Su, Tu, We etc. (\cite{BUI2017}). Corpora and annotation problems occur mainly due to merged words causing missing PHI, for example ``Hancevillewith''. Also, error is caused by false negative annotations (\cite{BUI2017}). Also, the reliance on using dictionaries and ML approaches, where unseen terms and unfamiliar contexts cause difficulties in identifying person and location resulted in errors. 

Dernoncourt \textit{et al.}, (2017) (\cite{doi:10.1093/jamia/ocw156}) were also interested in the following sources of the errors: abbreviations, ambiguities, data sparsity and debatable annotations. They found that abbreviations and ambiguities are the most challenging source of errors to address to improve the performance further. Abbreviations can be a source of confusion and error even for trained health professionals. For example, the dosing instruction ``QD'' can be interpreted as either ``once a day'' or ``four times a day''.  Sparsity may be resolved by increasing the size of the training set and including more instances of difficult PHI types. 

\section{Discussion}

This section presents a general discussion on de-identification of EHR followed by the summary and observations of the de-identification systems from the 2014 and 2016 competitions, and systems developed using competition data. Also, this section presents research questions for possible future avenues.  

\subsection{De-identification - a general discussion}

There is a great interest internationally in applying data-driven techniques to electronic health records (\cite{murdoch2013inevitable}). However, privacy laws in many jurisdictions
require accurate de-identification of medical documents, such as discharge summaries and electronic health records, before they can be shared outside of their originating institutions. De-identification of data provides opportunities for researchers to use medical data without individual consent. 

The sharing of records is crucial for advancing health research. For example, in the 2014 competition, the Heart Disease Risk Factors Challenge involved participating research groups attempting to predict heart disease risk factors in diabetic patients from longitudinal clinical narratives. As noted above, such a challenge would not have been possible under United States law if the narratives, 1,304 medical records from 296 diabetic patients, were not de-identified first. In this case, the records were de-identified with great care by multiple experts. Since most institutions would not be able to afford the costs of such de-identification, automating the process is therefore crucial for sharing data and advancing health research.

Accurate automatic de-identification systems also help gain public trust in accepting the use of medical data for research purposes. As noted above, patients may be very concerned about some elements of their records, such as sexual history or mental health being traced back to them (\cite{king2012perspectives}). Patient privacy needs to be protected by both the data holder and the data user (\cite{dilemmaRoop}). The data holder needs to ensure that the data is correctly de-identified before releasing it, and the data user should only request the data they need and use if for the purpose it was issued for to minimise the risk of re-identification. 

On the other hand, the existing HIPAA Safe Harbor standard may already limit researchers’ ability to perform various studies by masking relevant information, such as restricting location to three-digit zip codes (applicable to many social and public health studies), which may limit accuracy in health research (\cite{benitez2010beyond}). However, the Safe Harbor method has the advantage of being set by law, which reduces legal liability and reputation risk (\cite{dilemmaRoop,benitez2010beyond}).  

The alternative standard of de-identification allowed by HIPPA is Expert Determination, where a suitably qualified individual certifies that the record has been de-identified to the extent that the risk of re-identification is acceptably low. The level of de-identification needed under this standard depends on context and risk. This allows the data to have the maximum value for the research at hand while still maintaining privacy. However, Expert Determination requires enough people with the skills necessary to safely make these determinations, widely accepted industrial standards, and de-identification techniques that will meet these standards (\cite{benitez2010beyond,chester18}). 

\subsection{Summary of the competitions and  de-identification systems}

De-identification competitions such as that of 2014 and 2016 provide excellent opportunities for researchers to further develop research in this area. One primary reason these competitions are so widely recognised is the open source nature of the data. This paper provided an overview of 2014 and 2016 competitions. Due to the difference in the dataset and the content of the data, both competitions offered exciting developments in this area of research.  

In general, teams found the de-identification of 2016 dataset more difficult than that of 2014. Also, HIPAA PHI categories were more straightforward to de-identify than that of i2b2 PHI (for 2014 data) or N-GRID PHI (for 2016 data). As explained earlier, i2b2 PHI and N-GRID PHI include additional and more complex categories to the HIPAA regulations. The de-identification competitions argued that HIPAA PHI is not enough for complete de-identification of the EHR, especially for a longitudinal dataset. An example would be the detection of the difference between the patient's name and doctor's name. Although the de-identification of NAME has high F-measure, the differentiation mentioned above has not been achieved successfully. This adds an extra layer of complexity.

In the 2014 competition, top systems preferred hybrid approaches with CRF being the most used machine learning approach. Considering ten teams got very high F-measures (\cite{STUBBS2015S78}, it is evident that the gold standard had enough information presented. Lack of training examples may be one reason for the PROFESSION category performing so poorly among all systems presented in that competition (\cite{ STUBBS2015S11,YANG2015S30}). In the 2016 competition, as in 2014, most of the top systems were hybrid systems with CRF and/or neural networks systems combined with rule-based systems. In 2016, there was a noticeable shift in favouring deep learning techniques with hand corded rules for specific PHIs.

The top-performing system in the 2016 competitions achieved a level of accuracy very close to that recorded by the least experienced annotator (\cite{UZUNER2017S1}). However, it is important to note that 95\% rule of thumb for accuracy was not reached in either Task 1A or 1B (\cite{psychatric}). Although there was a greater emphasis on robustness, being able to customise a de-identification system to the data results in much better outcome (\cite{psych1}). Regarding protecting privacy the three categories that are harder to predict in 2016 competitions are also the ones that have the potential to disclose patient identity (\cite{psychatric}).

In both 2014 and 2016 competitions, the best approach was hybrid systems. It is clear that hybrid systems have the potential of using the best of both rules-based and machine learning-based approaches. Categories such as date and contact obtained high F-measure in recently developed systems. Both of these categories also used rules/regular expressions.  It is also important to point out that de-identification systems that were developed using the 2014 dataset, post competitions, favoured deep learning only systems, not hybrid systems.

There was a noticeable shift towards using neural networks in 2016 competitions and systems developed after the 2014 competitions using 2014 data. Deep learning techniques allow learning representations that are tailored to the task at hand, in this case, de-identification. These representations have shown to be stronger than hand-engineered features (\cite{goldberg2017neural}). This shift is welcome, but there is still more scope to achieve perfection, adaptability and reproducibility.

The results from both the 2014 and the 2016 competitions are very high for the top teams. However, it is important to note that there is a high variation in the frequency of occurrence of PHIs. For example, PHIs such as DATE, NAME and DOCTOR occur much more often than PHIs such as PROFESSION and ORGANISATION. These more frequently occurring PHIs are also considered to be the easier to find entities, compared to the less frequent PHIs. The higher F-measures seem to reflect this imbalance. Hence the results may be misleading, i.e. high average F-measure does not imply good de-identification performance on rare categories such as PROFESSION.  

It is also evident that compared to the systems presented in the 2014 competitions, 2016 competition systems are much more complicated. The resulting gain in accuracy seems rather small. This complexity makes it very hard to reproduce these systems and also make it hard for others to replicate their results for scientific purposes.

It is also important to point out that an expert manually developed most of the rule-based components of the systems for the given data. However, the adaptability of these systems to new data remains an open question. This was partly evident from track 1A 2016 competition results, where only a couple of systems produced average results on a new dataset (in this case 2016). Looking at transfer learning (\cite{Thrun97} is, therefore, an opportunity.  It might be worth exploring domains such as search engine logs and law data, in which de-identified data is less sensitive than it is in the medical field. Hence, it is possible to obtain large datasets suitable for transfer learning.

More competitions such as that of the 2014 and the 2016 de-identification competitions will undoubtedly provide more opportunities for solving the de-identification problem. Emphasis on tracks such as that of track 3 in the 2014 competition and track 1A in the 2016 competition is also vital. The need for simpler systems that can be reproduced and used in different applications is crucial.

One primary concern with these de-identification systems is how well they will perform with other datasets. Unfortunately, EHR differs in each institute, and sometimes even in each department (\cite{naturallang}). Will these systems adapt to the change in EHR? It is important to note both data for both 2006 and 2014 were drawn from the same source, though 2014 data was much more extensive than for 2006 (\cite{STUBBS2015S11}). However, the 2016 data was very different from that of the other competitions. As track 1A from 2016 competition was explicitly designed to address this issue, it was clear that apart from a couple of average results, the existing systems did not perform well when the dataset was changed (\cite{psychatric}). It is a good start for building models that can be tuned to new data, and with more competitions and new data, systems can be trained, and better accuracy might be achieved. 
 
These competitions have provided an excellent platform for developing and improving de-identification systems, but there is still a lot more work to be done to find a solution to the de-identification problem.

\subsection{Research Questions}

Although these de-identification competitions have provided a great platform to develop this area, some questions/areas are worth exploring. 

\begin{enumerate}
    \item Considering the noticeable changes across the systems - is the hybrid system the best way to go? Could machine learning based system (for example, LSTM) outperform and be more adaptable to a new dataset?
    \item Is the use of rules and regular expressions crucial for rare, occurring PHIs? If so can we learn rules and regular expressions from data instead of handwritten rules (as seen in these competitions)?
    \item Apart from the noted fact that some PHIs, such as PROFESSION and LOCATION, perform poorly due to the low occurrence in training data, is there any other reason for this? Especially considering the use of handwritten rules and dictionary in the competitions, what are some other possible issues related to the poor F-measure? Any possible solutions to these problems?
    \item Adaptability to a new dataset is a noticeable issue with the de-identification systems. This is partly due to the lack of datasets. Could transfer learning be used to solve this problem? If so, is there any similar datasets available from domains such as law and crime science?
    \item Is there a way to obtain similar results to that of 2016 competitions with less complicated, reproducible systems? 
    \item Considering the 95\% on HIPAA regulations is still to be achieved across all PHIs - is there a need for stricter regulations, such as that introduced in these competitions? If so the re-identification implications will be an interesting avenue to explore.
    \item Systems developed using 2014 datasets after the competitions reported better performances for poorly performing PHIs. Is there a reason for this change? Would these systems do equally well if used for a different dataset? Can techniques from these systems be adopted to develop a more robust and better performing de-identification system?
    \item Would approaches such as word embeddings, which try and model the semantics of words, better be able to capture rarely occurring PHIs as opposed to syntactic approaches (given the lack of examples in the datasets to model the syntax of the rare PHIs)?
\end{enumerate}

\section*{Acknowledgements}

We would like to acknowledge the New Zealand Precision Driven Health Research Partnership for funding the first author through a summer research scholarship.

\bibliographystyle{chicago}
\bibliography{bibliography_review}

\begin{thebibliography}{}

\bibitem[\protect\citeauthoryear{Aberdeen, Bayer, Yeniterzi, Wellner, Clark,
  Hanauer, Malin, and Hirschman}{Aberdeen et~al.}{2010}]{ABERDEEN2010849}
Aberdeen, J., S.~Bayer, R.~Yeniterzi, B.~Wellner, C.~Clark, D.~Hanauer,
  B.~Malin, and L.~Hirschman (2010).
\newblock The {MITRE} identification scrubber toolkit: Design, training, and
  assessment.
\newblock {\em International Journal of Medical Informatics\/}~{\em 79\/}(12),
  849 -- 859.

\bibitem[\protect\citeauthoryear{Benitez, Loukides, and Malin}{Benitez
  et~al.}{2010}]{benitez2010beyond}
Benitez, K., G.~Loukides, and B.~Malin (2010).
\newblock Beyond safe harbor: automatic discovery of health information
  de-identification policy alternatives.
\newblock In {\em Proceedings of the 1st ACM International Health Informatics
  Symposium}, pp.\  163--172. ACM.

\bibitem[\protect\citeauthoryear{Bui, Wyatt, and Cimino}{Bui
  et~al.}{2017}]{BUI2017}
Bui, D. D.~A., M.~Wyatt, and J.~J. Cimino (2017).
\newblock The {UAB} informatics institute and 2016 {CEGS N-GRID}
  de-identification shared task challenge.
\newblock {\em Journal of Biomedical Informatics\/}.

\bibitem[\protect\citeauthoryear{Cardinal}{Cardinal}{2017}]{CRATE}
Cardinal, R.~N. (2017).
\newblock Clinical records anonymisation and text extraction {(CRATE)}: an
  open-source software system.
\newblock {\em BMC Medical Informatics and Decision Making\/}~{\em 17\/}(1),
  50.

\bibitem[\protect\citeauthoryear{Chazard, Mouret, Ficheur, Schaffar, Beuscart,
  and Beuscart}{Chazard et~al.}{2014}]{CHAZARD2014303}
Chazard, E., C.~Mouret, G.~Ficheur, A.~Schaffar, J.-B. Beuscart, and
  R.~Beuscart (2014).
\newblock Proposal and evaluation of {FASDIM}, a fast and simple
  de-identification method for unstructured free-text clinical records.
\newblock {\em International Journal of Medical Informatics\/}~{\em 83\/}(4),
  303 -- 312.

\bibitem[\protect\citeauthoryear{Chen, Cullen, and Godwin}{Chen
  et~al.}{2015}]{CHEN2015S60}
Chen, T., R.~M. Cullen, and M.~Godwin (2015).
\newblock Hidden markov model using dirichlet process for de-identification.
\newblock {\em Journal of Biomedical Informatics\/}~{\em 58}, S60--S66.

\bibitem[\protect\citeauthoryear{Chester}{Chester}{2018a}]{iapp}
Chester, G. (2018a).
\newblock {International Association of Privacy Professionals}.
\newblock {https://iapp.org/news/}.
\newblock Accessed: July 09, 2018.

\bibitem[\protect\citeauthoryear{Chester}{Chester}{2018b}]{chester18}
Chester, G. (2018b).
\newblock On de-identification: why better standards and more experts are
  needed.
\newblock
  {https://iapp.org/news/a/on-de-identification-why-better-standards-and-more-experts-needed/}.
\newblock Last accessed: July 13, 2018.

\bibitem[\protect\citeauthoryear{Commissioner}{Commissioner}{2009}]{healthdisability}
Commissioner, H. .~D. (2009).
\newblock {The Code of Rights}, {Health \& Disability Commissioner}.
\newblock {http://www.hdc.org.nz/the-act--code/the-code-of-rights}.
\newblock Original date: 2009; Accessed: December 09, 2017.

\bibitem[\protect\citeauthoryear{Cormack, Nath, Milward, Raja, and
  Jonnalagadda}{Cormack et~al.}{2015}]{CORMACK2015S120}
Cormack, J., C.~Nath, D.~Milward, K.~Raja, and S.~R. Jonnalagadda (2015).
\newblock Agile text mining for the 2014 i2b2/uthealth cardiac risk factors
  challenge.
\newblock {\em Journal of Biomedical Informatics\/}~{\em 58\/}(Supplement),
  S120 -- S127.
\newblock Proceedings of the 2014 i2b2/UTHealth Shared-Tasks and Workshop on
  Challenges in Natural Language Processing for Clinical Data.

\bibitem[\protect\citeauthoryear{Dai, Shabbir, Chen, and Wu}{Dai
  et~al.}{2015}]{dai2015}
Dai, H., S.~A. Shabbir, C.-W. Chen, and C.-C. Wu (2015, 09).
\newblock Recognition and evaluation of clinical section headings in clinical
  documents using token-based formulation with conditional random fields.
\newblock {\em BioMed Research International\/}~{\em 2015}.

\bibitem[\protect\citeauthoryear{Dehghan, Kovacevic, Karystianis, Keane, and
  Nenadic}{Dehghan et~al.}{2015}]{DEHGHAN2015S53}
Dehghan, A., A.~Kovacevic, G.~Karystianis, J.~A. Keane, and G.~Nenadic (2015).
\newblock Combining knowledge- and data-driven methods for de-identification of
  clinical narratives.
\newblock {\em Journal of Biomedical Informatics\/}~{\em 58\/}(Supplement), S53
  -- S59.

\bibitem[\protect\citeauthoryear{Dehghan, Kovacevic, Karystianis, Keane, and
  Nenadic}{Dehghan et~al.}{2017}]{data-driven}
Dehghan, A., A.~Kovacevic, G.~Karystianis, J.~A. Keane, and G.~Nenadic (2017).
\newblock Learning to identify protected health information by integrating
  knowledge-and data-driven algorithms: A case study on psychiatric evaluation
  notes.
\newblock {\em Journal of Biomedical Informatics\/}~{\em 75}, S28--S33.

\bibitem[\protect\citeauthoryear{Deleger, Molnar, Savova, Xia, Lingren, Li,
  Marsolo, Jegga, Kaiser, Stoutenborough, and Solti}{Deleger
  et~al.}{2013}]{CCHMC}
Deleger, L., K.~Molnar, G.~Savova, F.~Xia, T.~Lingren, Q.~Li, K.~Marsolo,
  A.~Jegga, M.~Kaiser, L.~Stoutenborough, and I.~Solti (2013).
\newblock Large-scale evaluation of automated clinical note de-identification
  and its impact on information extraction.
\newblock {\em Journal of the American Medical Informatics Association\/}~{\em
  20\/}(1), 84--94.

\bibitem[\protect\citeauthoryear{Dernoncourt, Lee, and Szolovits}{Dernoncourt
  et~al.}{2017}]{DBLP:journals/corr/DernoncourtLS17}
Dernoncourt, F., J.~Y. Lee, and P.~Szolovits (2017).
\newblock Neuroner: an easy-to-use program for named-entity recognition based
  on neural networks.
\newblock {\em CoRR\/}~{\em abs/1705.05487}.

\bibitem[\protect\citeauthoryear{Dernoncourt, Lee, Uzuner, and
  Szolovits}{Dernoncourt et~al.}{2017}]{doi:10.1093/jamia/ocw156}
Dernoncourt, F., J.~Y. Lee, {\"O}.~Uzuner, and P.~Szolovits (2017).
\newblock De-identification of patient notes with recurrent neural networks.
\newblock {\em Journal of the American Medical Informatics Association\/}~{\em
  24\/}(3), 596--606.

\bibitem[\protect\citeauthoryear{Garfinkel}{Garfinkel}{2015}]{NIST}
Garfinkel, S. (2015).
\newblock De-identification of personally identifiable information technical
  report.
\newblock {\em National institute of Standards and Technology (NIST), U.S.
  Department of Commerce\/}.

\bibitem[\protect\citeauthoryear{Goldberg}{Goldberg}{2017}]{goldberg2017neural}
Goldberg, Y. (2017).
\newblock Neural network methods for natural language processing.
\newblock {\em Synthesis Lectures on Human Language Technologies\/}~{\em
  10\/}(1), 1--309.

\bibitem[\protect\citeauthoryear{He, Guan, Cheng, Cen, and Hua}{He
  et~al.}{2015}]{HE2015S39}
He, B., Y.~Guan, J.~Cheng, K.~Cen, and W.~Hua (2015).
\newblock {CRFs} based de-identification of medical records.
\newblock {\em Journal of Biomedical Informatics\/}~{\em 58\/}(Supplement), S39
  -- S46.

\bibitem[\protect\citeauthoryear{Jiang, Zhao, He, Guan, and Jiang}{Jiang
  et~al.}{2017}]{JIANG2017S43}
Jiang, Z., C.~Zhao, B.~He, Y.~Guan, and J.~Jiang (2017).
\newblock De-identification of medical records using conditional random fields
  and long short-term memory networks.
\newblock {\em Journal of Biomedical Informatics\/}~{\em 75\/}(Supplement), S43
  -- S53.

\bibitem[\protect\citeauthoryear{Khalifa and Meystre}{Khalifa and
  Meystre}{2015}]{KHALIFA2015S128}
Khalifa, A. and S.~Meystre (2015).
\newblock Adapting existing natural language processing resources for
  cardiovascular risk factors identification in clinical notes.
\newblock {\em Journal of Biomedical Informatics\/}~{\em 58\/}(Supplement),
  S128 -- S132.

\bibitem[\protect\citeauthoryear{King, Brankovic, and Gillard}{King
  et~al.}{2012}]{king2012perspectives}
King, T., L.~Brankovic, and P.~Gillard (2012).
\newblock Perspectives of australian adults about protecting the privacy of
  their health information in statistical databases.
\newblock {\em International Journal of Medical Informatics\/}~{\em 81\/}(4),
  279--289.

\bibitem[\protect\citeauthoryear{Kumar, Stubbs, Shaw, and Uzuner}{Kumar
  et~al.}{2015}]{KUMAR2015S6}
Kumar, V., A.~Stubbs, S.~Shaw, and {\"O}.~Uzuner (2015).
\newblock Creation of a new longitudinal corpus of clinical narratives.
\newblock {\em Journal of Biomedical Informatics\/}~{\em 58\/}(Supplement), S6
  -- S10.

\bibitem[\protect\citeauthoryear{Kushida, Nichols, Jadrnicek, Miller, Walsh,
  and Griffin}{Kushida et~al.}{2012}]{10.2307/23216664}
Kushida, C.~A., D.~A. Nichols, R.~Jadrnicek, R.~Miller, J.~K. Walsh, and
  K.~Griffin (2012).
\newblock Strategies for de-identification and anonymization of electronic
  health record data for use in multicenter research studies.
\newblock {\em Medical care\/}, S82--S101.

\bibitem[\protect\citeauthoryear{Lee, Wu, Zhang, Xu, Xu, and Roberts}{Lee
  et~al.}{2017}]{psych1}
Lee, H.-J., Y.~Wu, Y.~Zhang, J.~Xu, H.~Xu, and K.~Roberts (2017).
\newblock A hybrid approach to automatic de-identification of psychiatric
  notes.
\newblock {\em Journal of Biomedical Informatics\/}~{\em 75}, S19--S27.

\bibitem[\protect\citeauthoryear{Lee, Dernoncourt, and Szolovits}{Lee
  et~al.}{2017}]{DBLP:journals/corr/LeeDS17a}
Lee, J.~Y., F.~Dernoncourt, and P.~Szolovits (2017).
\newblock Transfer learning for named-entity recognition with neural networks.
\newblock {\em CoRR\/}~{\em abs/1705.06273}.

\bibitem[\protect\citeauthoryear{Li, Chai, Zhao, Nan, and Zhao}{Li
  et~al.}{2016}]{Li2016}
Li, K., Y.~Chai, H.~Zhao, X.~Nan, and Y.~Zhao (2016).
\newblock {\em Learning to Recognize Protected Health Information in Electronic
  Health Records with Recurrent Neural Network}, pp.\  575--582.
\newblock Cham: Springer International Publishing.

\bibitem[\protect\citeauthoryear{Liu, Chen, Tang, Wang, Chen, Li, Wang, Deng,
  and Zhu}{Liu et~al.}{2015}]{LIU2015S47}
Liu, Z., Y.~Chen, B.~Tang, X.~Wang, Q.~Chen, H.~Li, J.~Wang, Q.~Deng, and
  S.~Zhu (2015).
\newblock Automatic de-identification of electronic medical records using
  token-level and character-level conditional random fields.
\newblock {\em Journal of Biomedical Informatics\/}~{\em 58\/}(Supplement), S47
  -- S52.

\bibitem[\protect\citeauthoryear{Liu, Tang, Wang, and Chen}{Liu
  et~al.}{2017}]{LI2017}
Liu, Z., B.~Tang, X.~Wang, and Q.~Chen (2017).
\newblock De-identification of clinical notes via recurrent neural network and
  conditional random field.
\newblock {\em Journal of Biomedical Informatics\/}~{\em 75}, S34--S42.

\bibitem[\protect\citeauthoryear{Liu, Yang, Wang, Chen, Tang, Wang, and Xu}{Liu
  et~al.}{2017}]{Liu2017}
Liu, Z., M.~Yang, X.~Wang, Q.~Chen, B.~Tang, Z.~Wang, and H.~Xu (2017, Jul).
\newblock Entity recognition from clinical texts via recurrent neural network.
\newblock {\em BMC Medical Informatics and Decision Making\/}~{\em 17\/}(2),
  67.

\bibitem[\protect\citeauthoryear{Meystre, Friedlin, South, Shen, and
  Samore}{Meystre et~al.}{2010}]{Meystre2010}
Meystre, S., J.~Friedlin, B.~South, S.~Shen, and M.~Samore (2010, Aug).
\newblock Automatic de-identification of textual documents in the electronic
  health record: a review of recent research.
\newblock {\em BMC Medical Research Methodology\/}~{\em 10\/}(1), 70.

\bibitem[\protect\citeauthoryear{Meystre, Ferr{\'a}ndez, Friedlin, South, Shen,
  and Samore}{Meystre et~al.}{2014}]{MEYSTRE2014142}
Meystre, S.~M., {\'O}.~Ferr{\'a}ndez, F.~J. Friedlin, B.~R. South, S.~Shen, and
  M.~H. Samore (2014).
\newblock Text de-identification for privacy protection: A study of its impact
  on clinical text information content.
\newblock {\em Journal of Biomedical Informatics\/}~{\em 50\/}(Supplement C),
  142 -- 150.
\newblock Special Issue on Informatics Methods in Medical Privacy.

\bibitem[\protect\citeauthoryear{Murdoch and Detsky}{Murdoch and
  Detsky}{2013}]{murdoch2013inevitable}
Murdoch, T.~B. and A.~S. Detsky (2013).
\newblock The inevitable application of big data to health care.
\newblock {\em Jama\/}~{\em 309\/}(13), 1351--1352.

\bibitem[\protect\citeauthoryear{{Office of the Privacy Commissioner}}{{Office
  of the Privacy Commissioner}}{2013}]{healthprivacy}
{Office of the Privacy Commissioner} (2013).
\newblock {Health Information Privacy Code 1994.}
\newblock
  https://www.privacy.org.nz/the-privacy-act-and-codes/codes-of-practice/health-information-privacy-code-1994/.
\newblock Original date: 2013; Accessed: December 09, 2017.

\bibitem[\protect\citeauthoryear{Phuong and Chau}{Phuong and
  Chau}{2016}]{Phuong2016AutomaticDO}
Phuong, N.~D. and V.~T.~N. Chau (2016).
\newblock Automatic de-identification of medical records with a multilevel
  hybrid semi-supervised learning approach.
\newblock {\em 2016 IEEE RIVF International Conference on Computing \&
  Communication Technologies, Research, Innovation, and Vision for the Future
  (RIVF)\/}, 43--48.

\bibitem[\protect\citeauthoryear{Roop}{Roop}{2018}]{dilemmaRoop}
Roop, E. (2018).
\newblock The de-identification dilemma.
\newblock http://www.fortherecordmag.com/archives /0515p16.shtml.
\newblock Last accessed: June 17, 2018.

\bibitem[\protect\citeauthoryear{Rothstein}{Rothstein}{2010}]{rothstein2010deidentification}
Rothstein, M.~A. (2010).
\newblock Is deidentification sufficient to protect health privacy in research?
\newblock {\em The American Journal of Bioethics\/}~{\em 10\/}(9), 3--11.

\bibitem[\protect\citeauthoryear{Shweta, Kumar, Ekbal, Saha, and
  Bhattacharyya}{Shweta et~al.}{2016}]{kumar2016}
Shweta, A.~Kumar, A.~Ekbal, S.~Saha, and P.~Bhattacharyya (2016).
\newblock A recurrent neural network architecture for de-identifying clinical
  records.
\newblock In {\em Proceedings of the 13th Intl. Conference on Natural Language
  Processing}, pp.\  188--197.

\bibitem[\protect\citeauthoryear{Stubbs, Filannino, and Uzuner}{Stubbs
  et~al.}{2017}]{psychatric}
Stubbs, A., M.~Filannino, and {\"O}.~Uzuner (2017).
\newblock De-identification of psychiatric intake records: Overview of 2016
  {CEGS N-GRID} shared tasks track 1.
\newblock {\em Journal of Biomedical Informatics\/}.

\bibitem[\protect\citeauthoryear{Stubbs, Kotfila, and Uzuner}{Stubbs
  et~al.}{2015}]{STUBBS2015S11}
Stubbs, A., C.~Kotfila, and {\"O}.~Uzuner (2015).
\newblock Automated systems for the de-identification of longitudinal clinical
  narratives: Overview of 2014 {i2b2/UTHealth} shared task track 1.
\newblock {\em Journal of Biomedical Informatics\/}~{\em 58\/}(Supplement), S11
  -- S19.

\bibitem[\protect\citeauthoryear{Stubbs, Kotfila, Xu, and Uzuner}{Stubbs
  et~al.}{2015}]{STUBBS2015S67}
Stubbs, A., C.~Kotfila, H.~Xu, and {\"O}.~Uzuner (2015).
\newblock Identifying risk factors for heart disease over time: Overview of
  2014 {i2b2/UTHealth} shared task track 2.
\newblock {\em Journal of Biomedical Informatics\/}~{\em 58\/}(Supplement), S67
  -- S77.

\bibitem[\protect\citeauthoryear{Stubbs and Uzuner}{Stubbs and
  Uzuner}{2015a}]{STUBBS2015S20}
Stubbs, A. and {\"O}.~Uzuner (2015a).
\newblock Annotating longitudinal clinical narratives for de-identification:
  The 2014 {i2b2/UTHealth} corpus.
\newblock {\em Journal of Biomedical Informatics\/}~{\em 58\/}(Supplement), S20
  -- S29.

\bibitem[\protect\citeauthoryear{Stubbs and Uzuner}{Stubbs and
  Uzuner}{2015b}]{STUBBS2015S78}
Stubbs, A. and {\"O}.~Uzuner (2015b).
\newblock Annotating risk factors for heart disease in clinical narratives for
  diabetic patients.
\newblock {\em Journal of Biomedical Informatics\/}~{\em 58\/}(Supplement), S78
  -- S91.

\bibitem[\protect\citeauthoryear{Stubbs and Uzuner}{Stubbs and
  Uzuner}{2017}]{Stubbs2017}
Stubbs, A. and {\"O}.~Uzuner (2017).
\newblock De-identification of medical records through annotation.
\newblock In {\em Handbook of Linguistic Annotation}, pp.\  1433--1459.
  Springer.

\bibitem[\protect\citeauthoryear{Stubbs, Uzuner, Kotfila, Goldstein, and
  Szolovits}{Stubbs et~al.}{2015}]{Stubbs2015}
Stubbs, A., {\"O}.~Uzuner, C.~Kotfila, I.~Goldstein, and P.~Szolovits (2015).
\newblock {\em Challenges in Synthesizing Surrogate PHI in Narrative EMRs},
  pp.\  717--735.
\newblock Cham: Springer International Publishing.

\bibitem[\protect\citeauthoryear{Thrun and Pratt}{Thrun and
  Pratt}{1997}]{Thrun97}
Thrun, S. and L.~Pratt (1997).
\newblock {\em Learning To Learn}.
\newblock Norwell, MA: Kluwer Academic.

\bibitem[\protect\citeauthoryear{Tran and Kavuluru}{Tran and
  Kavuluru}{2017}]{TRAN2017S138}
Tran, T. and R.~Kavuluru (2017).
\newblock Predicting mental conditions based on “history of present
  illness” in psychiatric notes with deep neural networks.
\newblock {\em Journal of Biomedical Informatics\/}~{\em 75\/}(Supplement),
  S138 -- S148.

\bibitem[\protect\citeauthoryear{Uzuner, Luo, and Szolovits}{Uzuner
  et~al.}{2007}]{doi:10.1197/jamia.M2444}
Uzuner, {\"O}., Y.~Luo, and P.~Szolovits (2007).
\newblock Evaluating the state-of-the-art in automatic de-identification.
\newblock {\em Journal of the American Medical Informatics Association\/}~{\em
  14\/}(5), 550--563.

\bibitem[\protect\citeauthoryear{Uzuner and Stubbs}{Uzuner and
  Stubbs}{2015}]{naturallang}
Uzuner, {\"O}. and A.~Stubbs (2015).
\newblock Practical applications for natural language processing in clinical
  research: The 2014 i2b2/uthealth shared tasks.
\newblock {\em Journal of Biomedical Informatics\/}~{\em 58\/}(Suppl), S1.

\bibitem[\protect\citeauthoryear{Uzuner, Stubbs, and Filannino}{Uzuner
  et~al.}{2017}]{UZUNER2017S1}
Uzuner, {\"O}., A.~Stubbs, and M.~Filannino (2017).
\newblock A natural language processing challenge for clinical records:
  Research domains criteria {(RDoC)} for psychiatry.
\newblock {\em Journal of Biomedical Informatics\/}~{\em 75}, S1--S3.

\bibitem[\protect\citeauthoryear{Yang and Garibaldi}{Yang and
  Garibaldi}{2015}]{YANG2015S30}
Yang, H. and J.~M. Garibaldi (2015).
\newblock Automatic detection of protected health information from clinic
  narratives.
\newblock {\em Journal of Biomedical Informatics\/}~{\em 58\/}(Supplement), S30
  -- S38.

\bibitem[\protect\citeauthoryear{Zheng, Vydiswaran, Liu, Wang, Stubbs, Uzuner,
  Gururaj, Bayer, Aberdeen, Rumshisky, Pakhomov, Liu, and Xu}{Zheng
  et~al.}{2015}]{ZHENG2015S189}
Zheng, K., V.~V. Vydiswaran, Y.~Liu, Y.~Wang, A.~Stubbs, {\"O}.~Uzuner, A.~E.
  Gururaj, S.~Bayer, J.~Aberdeen, A.~Rumshisky, S.~Pakhomov, H.~Liu, and H.~Xu
  (2015).
\newblock Ease of adoption of clinical natural language processing software: An
  evaluation of five systems.
\newblock {\em Journal of Biomedical Informatics\/}~{\em 58\/}(Supplement),
  S189 -- S196.

\bibitem[\protect\citeauthoryear{Zuccon, Kotzur, Nguyen, and Bergheim}{Zuccon
  et~al.}{2014}]{DBLP:journals/artmed/ZucconKNB14}
Zuccon, G., D.~Kotzur, A.~N. Nguyen, and A.~Bergheim (2014).
\newblock De-identification of health records using anonym: Effectiveness and
  robustness across datasets.
\newblock {\em Artificial Intelligence in Medicine\/}~{\em 61\/}(3), 145--151.

\end{thebibliography}

\end{document}